\pdfoutput=1
\documentclass[11pt]{article}

\usepackage[]{ACL}

\usepackage{times}
\usepackage{latexsym}

\usepackage[T1]{fontenc}
\usepackage[utf8]{inputenc}

\usepackage{microtype}

\usepackage{prettyref}
\usepackage{xcolor}
\usepackage{rotating}
\usepackage{amsfonts,amssymb,amsmath,amsthm,amsopn,mathrsfs,mathtools,wasysym}	% math related
\usepackage{hhline,booktabs,colortbl,multirow,tabularx,diagbox,tablefootnote,threeparttable} % table related
\usepackage{subfig}
\usepackage{tikz}
\usepackage{makecell}
\usepackage{enumitem}

\usepackage{pifont}
\usepackage{algorithm}
\usepackage{algorithmic}

\usepackage{newfloat}
\usepackage{listings}

\newcommand{\cmark}{\ding{51}}%
\newcommand{\xmark}{\ding{55}}%

\def\sp1{\texttt{\textbf{Sentiment Cluster}}}
\def\sp2{\texttt{\textbf{Local Sentiment Coherency}}}

\def\our{\texttt{\textsc{LSA}}}
\def\ourx{\texttt{\textsc{LSA-X}}}

\def\ourt{\textcolor{blue}{\texttt{\textsc{LSA$_{T}$}}}}
\def\ours{\textcolor{green}{\texttt{\textsc{LSA$_{S}$}}}}
\def\ourp{\textcolor{magenta}{\texttt{\textsc{LSA$_{P}$}}}}
\def\oure{\textcolor{cyan}{\texttt{\textsc{LSA$_{E}$}}}}

\def\ourtx{\textcolor{blue}{\texttt{\textsc{LSA$_{T}$-X}}}}
\def\oursx{\textcolor{green}{\texttt{\textsc{LSA$_{S}$-X}}}}
\def\ourpx{\textcolor{magenta}{\texttt{\textsc{LSA$_{P}$-X}}}}
\def\ourex{\textcolor{cyan}{\texttt{\textsc{LSA$_{E}$-X}}}}

\newcommand{\pref}{\prettyref}

\newrefformat{fig}{Fig.~\ref{#1}}
\newrefformat{tab}{Table~\ref{#1}}
\newrefformat{sec}{Section~\ref{#1}}
\newrefformat{app}{Appendix~\ref{#1}}
\newrefformat{alg}{Algorithm~\ref{#1}}
\newrefformat{property}{Property~\ref{#1}}
\newrefformat{theorem}{Theorem~\ref{#1}}
\newrefformat{corollary}{Corollary~\ref{#1}}
\newrefformat{proposition}{Proposition~\ref{#1}}
\newrefformat{def}{Definition~\ref{#1}}
\newrefformat{eq}{equation~(\ref{#1})}

\title{Modeling Aspect Sentiment Coherency via Local Sentiment Aggregation}

\author{
	Heng Yang$^1$, Ke Li$^1$ \\ 
	$^1$Department of Computer Science, University of Exeter, EX4 4QF, Exeter, UK \\
	\texttt{\{hy345, k.li\}@exeter.ac.uk} 
}

\begin{document}
\maketitle
\begin{abstract}
	% Aspect sentiment coherency is an interesting but unexplored topic in the research of aspect-based sentiment classification (ABSC).
	% This concept reflects the common pattern where adjacent aspects often share similar sentiments.
	% Despite its prevalence, current studies haven't recognized the potential of modeling aspect sentiment coherency,
	% including its implications in adversarial defense. To model aspect sentiment coherency,
	% we propose a new local sentiment aggregation (\our) paradigm based on constructing a differential-weighted sentiment aggregation window.
	% One salient feature of our approach is its proficiency in extracting implicit sentiments associated with aspects lacking explicit sentiment descriptors.
	% We have rigorously evaluated our model through experiments, and the results attest to \our's proficiency in learning sentiment coherency.
	% It outperforms existing models and achieves state-of-the-art results across five public datasets.
	% Furthermore, we demonstrate the promising ability of \our\ in ABSC adversarial defense, thanks to sentiment coherency modeling.
	% To encourage further exploration and application of this concept, we have made our code publicly accessible.
	% This will provide researchers with a tool to explore sentiment coherency modeling in future research.

Aspect sentiment coherency is an intriguing yet underexplored topic in the field of aspect-based sentiment classification. This concept reflects the common pattern where adjacent aspects often share similar sentiments. Despite its prevalence, current studies have not fully recognized the potential of modeling aspect sentiment coherency, including its implications in adversarial defense. To model aspect sentiment coherency, we propose a novel local sentiment aggregation (\our) paradigm based on constructing a differential-weighted sentiment aggregation window. We have rigorously evaluated our model through experiments, and the results affirm the proficiency of \our\ in terms of aspect coherency prediction and aspect sentiment classification. For instance, it outperforms existing models and achieves state-of-the-art sentiment classification performance across five public datasets. Furthermore, we demonstrate the promising ability of \our\ in ABSC adversarial defense, thanks to its sentiment coherency modeling. To encourage further exploration and application of this concept, we have made our code publicly accessible. This will provide researchers with a valuable tool to delve into sentiment coherency modeling in future research.
\end{abstract}

\section{Introduction}
\label{intro}

\label{sec:sp}
\begin{figure*}[t!]
	\centering
	\includegraphics[width=\linewidth]{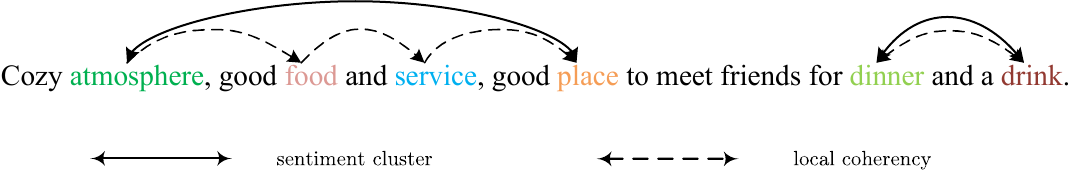}
	\caption{An example of aspect sentiment clusters and aspect sentiment coherency.}
	\label{fig:sp}
	\vspace{-10pt}
\end{figure*}

Aspect-based sentiment classification \cite{PontikiGPPAM14,PontikiGPMA15,PontikiGPAMAAZQ16} (ABSC) aims to identify sentiments associated with specific aspects within a text, as highlighted in several studies~\cite{MaLZW17, FanFZ18, ZhangLS19, YangZYSX21}. 
In this work, we make efforts to address an intriguing problem within ABSC that has been overlooked in existing research, i.e., ``\textbf{\textit{aspect sentiment coherency}}'', which focuses on modeling aspects that share similar sentiments. 
For instance, in the sentence ``\textit{This laptop has a lot of storage, and so does the battery capacity,}'' where ``\textit{storage}'' and ``\textit{battery capacity}'' aspects both contain positive sentiments. 
We show more examples of aspect sentiment coherency in \pref{fig:sp} and the case study section. 

The study of aspect sentiment coherency has not been investigated in existing research. 
Yet, some strides have been made on a similar topic, namely sentiment dependency. These approaches, featured in several studies~\cite{ZhangLS19, HuangC19a, PhanO20}, 
hypothesize that sentiments of aspects may be dependent and usually leverage syntax trees to reveal potential sentiment dependencies between aspects. 
However, sentiment dependency remains a somewhat ambiguous concept in the current research landscape. 
Furthermore, previous methods~\cite{ZhouHHH20, ZhaoHW20, TangJLZ20, LiCFMWH20, LiCFMWH20} tend to model context topological dependency (e.g., context syntax structure) rather than sentiment dependency directly. 
These techniques are resource-intensive and computation-intensive. Besides, they can suffer from token-node misalignment caused by conflicts in tokenization methods in syntax tree construction.

As a further contribution to current ABSC research, we propose aspect sentiment coherency learning and posit that modeling sentiment coherency can provide valuable insights. 
Modeling sentiment coherency often presents challenges for traditional ABSC methods due to the complexity of aspect sentiment coherency.
To efficiently address the aspect sentiment coherency task, we shed light on a simple yet effective approach, namely local sentiment aggregation (\our). 
More specifically, we introduce a local sentiment aggregation paradigm powered by three unique sentiment aggregation window strategies based on various aspect-based features to guide the modeling of aspect sentiment coherency. 
To comprehensively evaluate \our, we conduct experiments for the aspect sentiment coherency extraction subtask and the traditional aspect sentiment classification subtask. 
Our experimental results indicate that these strategies significantly enhance sentiment coherency modeling. 
\our\ achieves impressive performance in aspect sentiment coherency extraction and sentiment classification, setting new state-of-the-art results on five widely-used datasets based on the latest \texttt{DeBERTa}~\cite{HeGC21} model. Our work offers a new perspective on aspect-based sentiment analysis.

In conclusion, the main contributions of our work are as follows:
\begin{itemize}[nolistsep,noitemsep,leftmargin=*]
\item \textbf{Formulation}: We highlight the existence of sentiment coherency in ABSC and formulate the aspect sentiment coherency modeling task. Besides, we introduce a local sentiment aggregation mechanism to address this task.
\item \textbf{Method}: To implement the local sentiment aggregation mechanism, we introduce three strategies for constructing sentiment aggregation windows, demonstrating the effectiveness of our model in sentiment coherency modeling. 
We enhance this mechanism through differential weighted sentiment aggregation, allowing for dynamic adjustment of the aggregation window construction. 
\item \textbf{Evaluation}: According to our extensive experimental results, \our\ achieve impressive aspect sentiment coherency prediction results.
 Besides, our ensemble \our\ model also obtains state-of-the-art aspect sentiment classification performance on five public datasets. 
\end{itemize}
The codes and datasets related to this work are open-sourced at \url{https://github.com/yangheng95/PyABSA}.

\section{Sentiment Coherency}

We first introduce the concept of sentiment coherency and then formulate two sentiment coherency patterns.
In the review about a restaurant in \pref{fig:sp}, the reviewer expresses positive sentiments about the atmosphere, food, and service but remains neutral about dinner and drinks.
This tendency to express similar sentiments about related aspects (e.g., atmosphere, food, and service) is what we refer to as \textbf{\textit{sentiment coherency}}.
We calculate the number of sentiment clusters across all experimental datasets to prove this is a common phenomenon. The statistics are available in \pref{tab:cluster}.

Our aim is to study the extraction of aspect sentiment coherency and the improvement of ABSC performance by incorporating sentiment coherency.
We formulate two sentiment coherency patterns in the following sections.

\subsection{Aspect Sentiment Clusters}
Consider the example in \pref{fig:sp}. We notice that similar sentiments about different aspects tend to stick together, which is called \textbf{\textit{sentiment cluster}}.
The formulation of aspect sentiment clusters is as follows:
\begin{equation}
	\mathcal{C}=\{C_i \mid C_i = \{a_{1}, a_{2}, \ldots, a_{j} \}\},
\end{equation}
where $C_i$ is the $i$-th aspect sentiment cluster and $a_{j}$ is the $j$-th aspect in $C_i$, $1 \leq j \leq m$. $m$ is the number of identified aspects in the sentence.
Aspect sentiment clustering aims at concurrently predicting all sentiment clusters based on the provided aspects.
Aspect sentiment clusters can be regarded as a coarse-grained manifestation of sentiment coherency.
However, directly extracting these clusters can be quite challenging. We explain the challenges in the \pref{app:challenges}.
In consequence, we focus on asynchronous sentiment cluster prediction based on local sentiment coherency.

\subsection{Local Sentiment Coherency}
We propose \textit{``local coherency''} to simplify the modeling of aspect sentiment cluster extraction.
Local coherency utilizes the aspect features to predict the sentiment iteratively.
Finally, the aspects with the same sentiments are aggregated to predict sentiment clusters.
There are two advantages of local sentiment coherency modeling.
First, it helps us infer the sentiment about an aspect even when it isn't explicitly stated
(e.g., deriving that the reviewer had a positive dining experience without saying it outright).
Second, it smooths out the sentiment predictions, reducing errors caused by random noise or adversarial attacks.
As a result, we can have a more accurate understanding of sentiments.

\begin{table}[htbp]
	\centering
	\caption{The statistics of aspect sentiment clusters. "Cluster size" indicates the number of aspects in clusters with different sizes. }
	\resizebox{\linewidth}{!}{
		\begin{tabular}{l|c|c|c|c|c|c}
			\hline
			\multirow{2}[1]{*}{\textbf{Dataset}} & \multicolumn{5}{c|}{\textbf{Cluster Size}} & \multirow{1}[4]{*}{\textbf{Sum}}                                       \\
			\cline{2-6}                          & $1$                                        & $2$                              & $3$    & $4$   & $\geq 5$ &         \\
			\hline
			\texttt{Laptop14}                  & $791$                                      & $799$                            & $468$  & $294$ & $614$    & $2966$  \\
			\hline
			\texttt{Restaurant14}                    & $1318$                                     & $1050$                           & $667$  & $479$ & $1214$   & $4728$  \\
			\hline
			\texttt{Restaurant15}                    & $617$                                      & $406$                            & $229$  & $163$ & $326$    & $1741$  \\
			\hline
			\texttt{Restaurant16}                    & $836$                                      & $539$                            & $314$  & $210$ & $462$    & $2361$  \\
			\hline
			\texttt{MAMS}                        & $6463$                                     & $2583$                           & $1328$ & $746$ & $1397$   & $12517$ \\
			\hline
		\end{tabular}%
	}
	\label{tab:cluster}%
\end{table}%

\section{Methodology}
\label{sec:method}
In this section, we propose a local sentiment aggregation method for sentiment cluster prediction, which is based on the local sentiment coherency pattern.
We first introduce the implementation of local sentiment aggregation, which is based on sentiment window aggregation.
Then, we present the aspect feature learning method used for sentiment aggregation window construction in \pref{sec:saw}.
Finally, we describe the implementation details of our model.

\subsection{Local Sentiment Aggregation}
\label{sec:lsa}
To leverage local sentiment coherency, we extract the local sentiment information of each aspect and build a sentiment aggregation window
(which will be clarified in \pref{sec:saw}) to aggregate coherent sentiments.
In essence, the sentiment aggregation window is created by concatenating the feature representation of the aspect's local sentiment information
(i.e., aspect feature in the following sections). We propose three variants, \ourp, \ourt, and \ours, to construct sentiment aggregation windows.
\pref{fig:lsa-spc} illustrates the architecture of \ourp, while \pref{fig:lsa-lcf} presents the architecture of both \ourt\ and \ours.
The difference between \ourt\ and \ours\ is in the aspect feature used for local sentiment aggregation.

\subsection{Aspect Feature Learning}
\label{sec:saw}

Inspired by the existing studies, we employ the following aspect feature representations for local sentiment aggregation:
\begin{itemize}[leftmargin=*,noitemsep,nolistsep]
	\item Sentence pair-based (BERT-SPC) aspect feature~\cite{DevlinCLT19} (employed in \ourp)
	\item Local context focus-based (LCF) aspect feature~\cite{YangZYSX21} (employed in \ourt)
	\item Syntactical LCF-based (LCFS) based aspect feature~\cite{PhanO20} (employed in \ours)
\end{itemize}
We also present an ensemble model (\oure) that make use of the three variants of aspect-specific features.

\subsubsection{Sentence Pair-based Aspect Feature}

A straightforward way to obtain aspect features is to utilize the BERT-SPC input format~\cite{DevlinCLT19},
which appends the aspect to the context to learn aspect features.
For example, let $\mathcal{W} =\left\{[CLS], \{w^c_{i}\}_{i=1}^{n}, [SEP], \{w^a_{j}\}_{j=1}^{m}, [SEP]\right\}$ be the BERT-SPC format input,
$i \in [1, n]$ and $j \in [1, m]$, where $w^c_{i}$ and $w^a_{j}$ denote the token in the context and the aspect, respectively.
A PLM (e.g., BERT) can learn the aspect feature because the duplicated aspects will get more attention in the self-attention mechanism~\cite{VaswaniSPUJGKP17}.
As it is shown in \pref{fig:lsa-spc}, we simply apply the sentiment aggregation to BERT-SPC-based aspect features.
Note that we deploy a self-attention encoder before each linear layer to activate hidden states. 
We show the architecture of \ourp\ in \pref{fig:lsa-spc}.

\subsubsection{Local Context-based Aspect Feature}

\begin{figure}[htbp]
	\centering
	\includegraphics[width=\columnwidth]{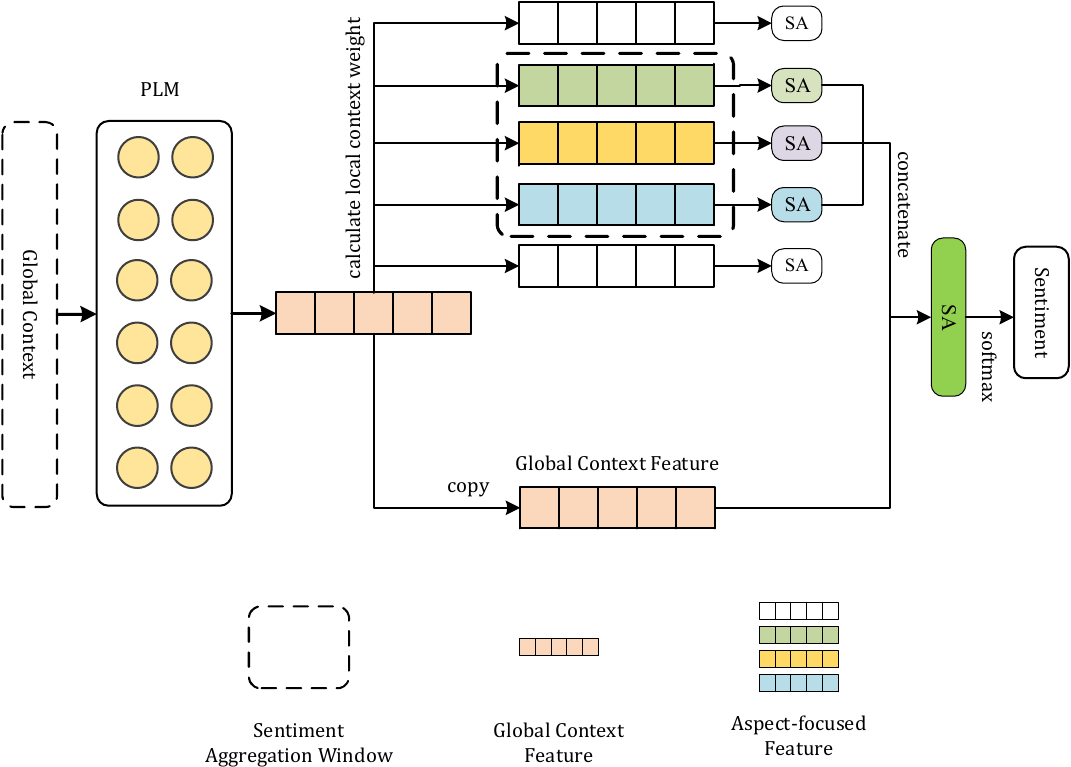}
	\caption{The local sentiment aggregation paradigm based on LCF/LCFS, denoted as \ourt\ and \ours.}
	\label{fig:lsa-lcf}
	\vspace{-10pt}
\end{figure}

The second implementation of our model is referred to as \ourt.
The local context-based aspect feature is derived by position-wise weighting the global context feature,
where the weights are calculated using the relative distance of token-aspect pairs.
Let $\mathcal{W} = \{w^c_{1}, w^c_{2}, \ldots, w^c_{n}\}$ be the tokens after tokenization.
We calculate the position weight for token $w^c_{i}$ as follows:
\begin{equation}
	\label{eq:1}
	\mathbf{H}_{w^{c}_{i}}^{*}:=\left\{\begin{array}{cc}
		\mathbf{H}^{c}_{w^c_{i}}                                                   & d_{w^c_{i}} \leq \alpha \\
		1-\frac{\left(d_{w^c_{i}}-\alpha\right)}{n} \cdot \mathbf{H}^{c}_{w^c_{i}} & d_{w^c_{i}} >\alpha
	\end{array}\right.,
\end{equation}
where $\mathbf{H}^{*}_{w^c{i}}$ and $\mathbf{H}^{c}_{w^c{i}}$, $i \in [1, n]$, are the hidden states at the position of $w^c_{i}$ in the aspect feature and global context feature, respectively.
$d_{w^c_{i}}$ is the relative distance between $w^c_{i}$ and the aspect. We concatenate $\mathbf{H}_{w^{c}_{i}}^{*}$ to obtain the aspect feature $\mathbf{H}_{}^{*}$.
$\alpha=3$ is a fixed distance threshold. If $d_{w^c_{i}} \leq\alpha$, $\mathbf{H}^{c}_{w^c{i}}$ will be preserved; otherwise, it decays according to $d_{w^c_{i}}$.

In \pref{eq:1}, the relative distance $d_{w^c_{i}}$ between $w^c_{i}$ and the aspect is obtained by:
\begin{equation}
	d_{w^c_{i}} := \frac{\sum_{j=1}^{m} |p^{c}_{i}-p^{a}_{j}|}{m},
\end{equation}
where $p^c_{i}$ and $p^{a}_{j}$ are the positions of the $w^c{i}$ and $j$-th token in the aspect.
As shown in \pref{fig:lsa-lcf}, we take the global context feature as a supplementary feature to learn aspect sentiments.

\subsubsection{Syntactical Local Context-based Aspect Feature}
The final variant of our model is \ours, which adopts the syntax-tree-based local context feature to construct a sentiment aggregation window.
The distance between the context word $w^c_{i}$ and the aspect can be calculated according to the shortest node distance between $w^c_{i}$ and the aspect in the syntax tree.
To leverage the syntactical information without directly modeling the syntax tree, \ours\ calculates the average node distance between $w^c_{i}$ and the aspect:
\begin{equation}
	d_{w^c_{i}} = \frac{\sum_{i=j}^{m} dist(w^{c}_{i}, w^{a}_{j})}{m},
\end{equation}
where $dist$ denotes the shortest distance between the node of $w^{c}_{i}$ and the node of $w^{a}_{j}$ in the syntax tree; the calculation of $\mathbf{H}_{w^{c}_{i}}^{*}$ follows \ourt.

\subsection{Sentiment Aggregation Window}
The sentiment aggregation window consists of $k$-nearest aspect feature vectors. Given that most of the clusters are small, we only consider $k=1$ in this study:
\begin{equation}
	\mathbf{H}_{aw}^{o} := [\{\mathbf{H}_{k}^{\mathbf{l}}\}; \mathbf{H}_{}^{\mathbf{t}}; \{\mathbf{H}_{k}^{\mathbf{r}}\}],
\end{equation}
\begin{equation}
	\mathbf{H}^{o} := W^{o} \mathbf{H}_{aw}^{o} + b^{o},
\end{equation}
where $\mathbf{H}_{aw}^{o}$ is the feature representation learned by local sentiment aggregation;
";" denotes vector concatenation. ${\mathbf{H}_{k}^{\mathbf{l}}}$ and ${\mathbf{H}_{k}^{\mathbf{r}}}$ are the $k$ nearest left and right adjacent aspect features, respectively.
$\mathbf{H}_{*}^{\mathbf{t}}$ is the targeted aspect feature. $\mathbf{H}_{*}^{o}$ is the representation learned by the sentiment aggregation window, and $W^{o}$ and $b^{o}$ are the trainable weights and biases.

\subsubsection{Aggregation Window Padding}

\begin{figure}[ht]
	\centering
	\includegraphics[width=\columnwidth]{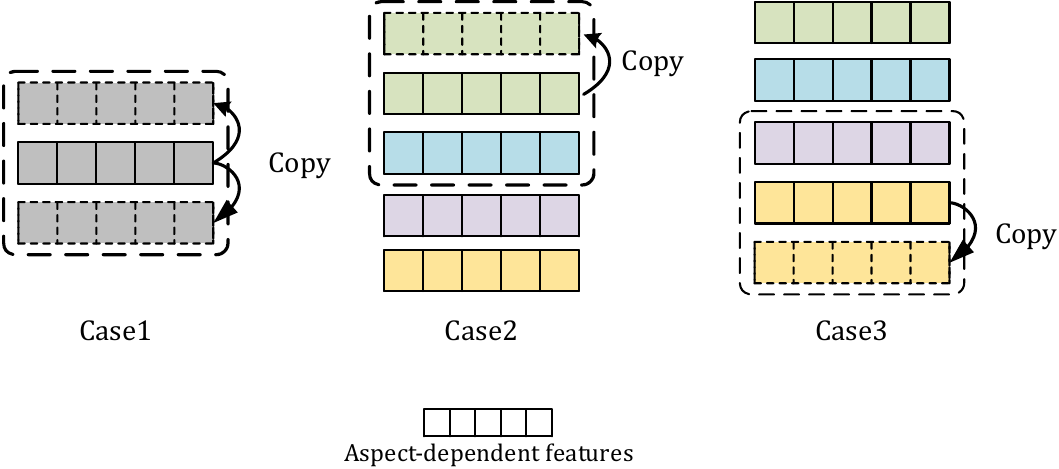}
	\caption{Window padding strategies for different situations. }
	\label{fig:padding}
	\vspace*{-10pt}
\end{figure}

To handle instances with no adjacent aspects, we pad the sentiment aggregation window. \pref{fig:padding} illustrates three padding strategies.
Instead of zero vectors, we pad the window using the targeted aspect's feature to highlight the local sentiment feature of the targeted aspect and prevent the model's performance from deteriorating.
Case \#$1$ indicates a single aspect in the context, in which we triple the targeted aspect's feature to build the sentiment aggregation window.
Case \#$2$ and Case \#$3$ duplicate the targeted aspect's feature to the left and right slots in the window, respectively.

\subsubsection{Differential Weighted Aggregation}
It is reasonable to assume that the importance of sentiment information from different sides may vary.
Therefore, we introduce differential weighted aggregation (DWA) to control the contribution of sentiment information from the adjacent aspects on different sides.
We initialize learnable $\eta_{l}^{*}$ and $\eta_{r}^{*}$ to $1$ and optimize them using gradient descent. The differential weighted sentiment aggregation window is obtained as follows:
\begin{equation}
	\mathbf{H}_{dwa}^{o} := [\eta_{l}^{*}\{\mathbf{H}_{k}^{\mathbf{l}}\}; \mathbf{H}_{}^{\mathbf{t}}; \eta_{r}^{*}\{\mathbf{H}_{k}^{\mathbf{r}}\}],
\end{equation}
where $\mathbf{H}_{dwa}^{o}$ is the aggregated hidden state learned by the differential weighted aggregation window.

\subsection{Output Layer}
For sentence pair-based sentiment aggregation, we simply apply pooling and softmax to predict the sentiment likelihood.
For the local context feature-based sentiment aggregation, we adhere to the original approach of combining the global context feature and the learned feature to predict sentiment polarity as follows:
\begin{equation}
	\mathbf{H}^{out} := W^{d}[\mathbf{H}^{o}; \mathbf{H}^{c}] + b^{d},
\end{equation}
where $\mathbf{H}^{out}$ is the output hidden state; $\mathbf{H}^{o}$ and $\mathbf{H}^{c}$ are the features extracted by a PLM (e.g., \texttt{DeBERTa}).
We use the feature of the first token (also known as the head pooling) to classify sentiments:
\begin{equation}
	\hat{y} :=\frac{\exp (\mathbf{h}^{head})}{\sum_{1}^{\Tilde{C}} \exp (\mathbf{h}^{head})},
\end{equation}
where $\mathbf{h}^{head}$ is the head-pooled feature; $\Tilde{C}$ is the number of polarity categories.
$W^{d} \in \mathbb{R}^{1 \times \Tilde{C}}$, $b^{d} \in \mathbb{R}^{\Tilde{C}}$ are the trainable weights and biases. $\hat{y}$ is the predicted sentiment polarity.

\subsection{Training Details}

The variants of our model based on different PLMs are denoted as \our-\texttt{BERT}, \our-\texttt{RoBERTa}, \our-\texttt{DeBERTa}, etc. \ourx-\texttt{DeBERTa} represents our model based on the large version of PLM\footnote{\url{https://huggingface.co/microsoft/deberta-v3-large}}.

We train our model using the AdamW optimizer with the cross-entropy loss function:
\begin{equation}
	\mathcal{L}=-\sum_{1}^{\Tilde{C}} \widehat{y_{i}} \log y_{i} + \lambda||\Theta||_{2} + \lambda^{*}||{\eta_{l}^{*},\eta_{r}^{*}}||_{2},
\end{equation}
where $\lambda$ is the $L_{2}$ regularization parameter; $\Theta$ is the parameter set of the model.
As we employ gradient-based optimization for $\eta_{l}^{*}$ and $\eta_{r}^{*}$, we also apply a $L_{2}$ regularization with $\lambda^{*}$ for $\eta_{l}^{*}$ and $\eta_{r}^{*}$.

\section{Experiments}
In this section, we introduce the settings of our experiments and report the experimental results.
We report all implementation details in the appendix, e.g., hyperparameter settings (\pref{app:hyperparameters}), baseline introduction (\pref{app:compared_models}) and additional experiments, etc.

\subsection{Datasets}
\label{app:datasets}
To evaluate the efficacy of the local sentiment aggregation, we conducted experiments on five popular ABSC datasets
\footnote{We evaluate \our\ on the \texttt{Twitter}~\cite{DongWTTZX14} dataset and report the experimental results in \pref{sec:twitter}.
  The processed datasets are available with the code in supplementary materials.}: \texttt{Laptop14}, \texttt{Restaurant14}, \texttt{Restaurant15} and \texttt{Restaurant16} datasets,
and \texttt{MAMS} dataset~\cite{JiangCXAY19}, respectively. The statistics of these datasets are shown in \pref{tab:datasets}.
\begin{table}[htb]
  \centering
  \caption{The statistics of all datasets used in our experiments. Note that in our experiments, only the MAMS dataset has a validation set.}
  \resizebox{\linewidth}{!}{
    \begin{tabular}{l|c|c|c|c|c|c}
      \hline
      \multirow{1}[4]{*}{\textbf{Datasets}} & \multicolumn{2}{c|}{\textbf{Positive}} & \multicolumn{2}{c|}{\textbf{Negative}} & \multicolumn{2}{c}{\textbf{Neutral}}                          \\
      \cline{2-7}                           & Train                         & Test                          & Train                       & Test  & Train  & Test  \\
      \hline
      \texttt{Laptop14}                     & $994$                         & $341$                         & $870$                       & $128$ & $464$  & $169$ \\
      \hline
      \texttt{Restaurant14}                       & $2164$                        & $728$                         & $807$                       & $196$ & $637$  & $196$ \\
      \hline
      \texttt{Restaurant15}                       & $909$                         & $326$                         & $256$                       & $180$ & $36$   & $34$  \\
      \hline
      \texttt{Restaurant16}                       & $1240$                        & $468$                         & $437$                       & $117$ & $69$   & $30$  \\
      \hline
      \texttt{MAMS}                         & $3379$                        & $400$                         & $2763$                      & $329$ & $5039$ & $607$ \\
      \hline
    \end{tabular}%
  }
  \label{tab:datasets}%
\end{table}%

\subsection{Hyperparameter Settings }
We introduce the hyperparameter settings in fine-tuning experiments.
\label{app:hyperparameters}
\begin{itemize}[leftmargin=*,noitemsep,nolistsep]
	\item We set $k=1$ in sentiment aggregation window construction.
	\item The learning rate for pre-trained models (e.g., \texttt{BERT} and \texttt{DeBERTa}) is $2\times10^{-5}$.
	\item The learning rates for $\eta_{l}^{*}$ and $\eta_{r}^{*}$ are both $0.01$.
	\item The batch size and maximum text modeling length are $16$ and $80$, respectively.
	\item The $L_2$ regularization parameters $\lambda$ and $\lambda_{*}$ are both $10^{-5}$.
\end{itemize}
We conduct experiments based on multiple PLMs. We implement our model based on the transformers: \url{https://github.com/huggingface/transformers}.

% \subsection{Evaluation Metrics}
% \label{app:metrics}
% We use the following metrics to evaluate the performance of \our.
% \begin{itemize}[leftmargin=*,noitemsep,nolistsep]
%   \item \textbf{Exact Match} (\texttt{EM}) is a metric that measures the performance of aspect sentiment cluster prediction (exactly matched cluster).
%   \item \textbf{Accuracy} (\texttt{Acc}) is a traditional metric to evaluate the performance of aspect sentiment classification.
%   \item \textbf{Macro \texttt{F$1$}} (\texttt{F$1$}) is another traditional metric which aims at alleviating the class imbalance problem in evaluation.
% \end{itemize}

\subsection{Baselines}
\label{app:compared_models}
In our comparative analysis, we evaluate the performance of \our\ in relation to several state-of-the-art ABSC models, many of which are syntax-based methods. 
These models include \texttt{SK-GCN-BERT}~\cite{ZhouHHH20}, which utilizes graph convolutional networks (GCN) to incorporate syntax and commonsense information for sentiment learning. 
\texttt{DGEDT-BERT}~\cite{TangJLZ20} is a dual-transformer-based network enhanced by a dependency graph, while \texttt{SDGCN-BERT}~\cite{ZhaoHW20} is a GCN-based model designed to capture sentiment dependencies between aspects. 
\texttt{Dual-GCN}~\cite{LiCFMWH20} is an innovative GCN-based model that enhances the learning of syntax and semantic features.

Additionally, we include models improved by \citet{DaiYSLQ21}, such as \texttt{RGAT-RoBERTa}, \texttt{PWCN-RoBERTa}, and \texttt{ASGCN-RoBERTa}, 
which leverage \texttt{RoBERTa} to induce syntax trees that align with \texttt{RoBERTa}'s tokenization strategy. 
\texttt{TGCN-BERT}~\cite{TianCS21} introduces a type-aware GCN that uses an attention mechanism to measure the importance of each edge in the syntax structure graph. 
\texttt{SARL-RoBERTa}~\cite{WangSLZC21} employs adversarial training to mitigate sentiment bias and align aspects with opinion words using span-based dependency. 
Finally, \texttt{dotGCN-BERT}~\cite{ChenTW022}, \texttt{SSEGCN-BERT}~\cite{ZhangZW22}, and \texttt{TGCN-BERT}~\cite{LiCFMWH20} are also included in our comparison. 
These models represent the current landscape of ABSC research, allowing us to assess the effectiveness of \our\ against well-established approaches.

We do not compare with \citet{CaoLPJB22} because we fail to find the source code of their model.

\subsection{Main Results}
\label{app:main_results}

We report sentiment coherency modeling performance and sentiment classification performance in this section.
\begin{table}[htbp]
  \centering
  \caption{The exact match score (\texttt{EM}) of sentiment cluster prediction on five public datasets The best results are highlighted in \textbf{bold} font. \texttt{Rest14}, \texttt{Rest15} and \texttt{Rest16} indicate \texttt{Restaurant14}, \texttt{Restaurant15} and \texttt{Restaurant16}, respectively.}
  \resizebox{\linewidth}{!}{
    \begin{tabular}{l|c|c|c|c|c}
      \hline
      \multicolumn{1}{l|}{\multirow{2}[1]{*}{\textbf{Model}}} & \texttt{Laptop$14$} & \texttt{Rest14}  & \texttt{Rest15}  & \texttt{Rest16}  & \texttt{MAMS}    \\
      \cline{2-6}  \multicolumn{1}{c|}{\multirow{2}[1]{*}{}}  & \texttt{EM}         & \texttt{EM}      & \texttt{EM}      & \texttt{EM}      & \texttt{EM}      \\
      \hline
      \texttt{BERT}                                           & $75.08$             & $78.75$          & $80.00$          & $87.60$          & $79.26$          \\
      \texttt{DeBERTa}                                        & $79.61$             & $83.88$          & $84.05$          & $89.72$          & $81.16$          \\
      % \texttt{SARL-BERT}                                     &                     &                 &                 &                 &               \\
      % \texttt{ASGCN-BERT}                                    &                     &                 &                 &                 &               \\
      \hline
      \texttt{\ourp-BERT}                                     & $78.14$             & $82.24$          & $82.76$          & $88.96$          & $82.35$          \\
      \texttt{\ourt-BERT}                                     & $78.06$             & $82.96$          & $82.66$          & $90.02$          & $82.46$          \\
      \texttt{\ours-BERT}                                     & $78.63$             & $83.09$          & $83.30$          & $88.75$          & $82.73$          \\
      \texttt{\oure-BERT}                                     & $78.94$             & $83.62$          & $83.40$          & $89.96$          & $84.03$          \\
      \hline
      \texttt{\ourp-DeBERTa}                                  & $82.55$             & $86.39$          & $86.93$          & $92.14$          & $82.83$          \\
      \texttt{\ourt-DeBERTa}                                  & $81.96$             & $86.26$          & $87.03$          & $91.72$          & $83.38$          \\
      \texttt{\ours-DeBERTa}                                  & $82.94$             & $85.90$          & $87.13$          & $91.87$          & $83.92$          \\
      \texttt{\oure-DeBERTa}                                  & $\mathbf{83.73}$    & $\mathbf{86.53}$ & $\mathbf{87.91}$ & $\mathbf{92.57}$ & $\mathbf{84.12}$ \\
      \hline
    \end{tabular}%
  }
  \label{tab:cluster_prediction}%
  \vspace{-10pt}
\end{table}%

\begin{table*}[htbp]
  \centering
  %   \scriptsize
  \caption{The traditional aspect sentiment classification performance on five public datasets, and the best results are heightened in \textbf{bold} font.
    $^{\dagger}$ indicates the results are the best performance in multiple runs, while other methods report the average performance.
    $^{\ddagger}$ indicates the experimental results of the models implemented by us. }
  \resizebox{\linewidth}{!}{
    \begin{tabular}{l|c|c|c|c|c|c|c|c|c|c|c}
      \hline
      \multicolumn{1}{l}{\multirow{2}[1]{*}{\textbf{Model}}} &                                                              & \multicolumn{2}{c|}{\texttt{Laptop$14$}} & \multicolumn{2}{c|}{\texttt{Restaurant14}} & \multicolumn{2}{c|}{\texttt{Restaurant15}} & \multicolumn{2}{c|}{\texttt{Restaurant16}} & \multicolumn{2}{c}{\texttt{MAMS}}                                                                                                                              \\
      \cline{3-12}  \multicolumn{1}{c}{\multirow{2}[1]{*}{}} &                                                              & \texttt{Acc}                             & \texttt{F$1$}                              & \texttt{Acc}                               & \texttt{F$1$}                              & \texttt{Acc}                      & \texttt{F$1$}          & \texttt{Acc}           & \texttt{F$1$}          & \texttt{Acc}           & \texttt{F$1$}          \\

      \hline

      \texttt{SK-GCN-BERT}~\cite{ZhouHHH20}                  & \multirow{14}[7]{*}{\begin{sideways}Baselines\end{sideways}} & $79.00$                                  & $75.57$                                    & $83.48$                                    & $75.19$                                    & $83.20$                           & $66.78$                & $87.19$                & $72.02$                & ---                    & ---                    \\
      \texttt{SDGCN-BERT}~\cite{ZhaoHW20}                    &                                                              & $81.35$                                  & $78.34$                                    & $83.57$                                    & $76.47$                                    & ---                               & ---                    & ---                    & ---                    & ---                    & ---                    \\
      \texttt{DGEDT-BERT}~\cite{TangJLZ20}                   &                                                              & $79.80$                                  & $75.60$                                    & $86.30$                                    & $80.00$                                    & $ 84.00$                          & $71.00$                & $91.90$                & $79.00$                & ---                    & ---                    \\
      \texttt{DualGCN-BERT}~\cite{LiCFMWH20}                 &                                                              & $81.80$                                  & $78.10$                                    & $87.13$                                    & $81.16$                                    & ---                               & ---                    & ---                    & ---                    & ---                    & ---                    \\
      \texttt{TF-BERT}~\cite{hangZLBWSX23}                   &                                                              & $81.80$                                  & $78.46$                                    & $87.09$                                    & $81.15$                                    & ---                               & ---                    & ---                    & ---                    & ---                    & ---                    \\
      \texttt{dotGCN-BERT}~\cite{ChenTW022}                  &                                                              & $81.03$                                  & $78.10$                                    & $86.16$                                    & $80.49$                                    & ---                               & ---                    & ---                    & ---                    & ---                    & ---                    \\
      \texttt{SSEGCN-BERT}~\cite{ZhangZW22}                  &                                                              & $81.01$                                  & $77.96$                                    & $87.31$                                    & $81.09$                                    & ---                               & ---                    & ---                    & ---                    & ---                    & ---                    \\
      \texttt{TGCN-BERT}~\cite{LiCFMWH20}                    &                                                              & $80.88$                                  & $77.03$                                    & $86.16$                                    & $79.95$                                    & $83.38$                           & $82.77$                & $86.00$                & $72.81$                & ---                    & ---                    \\
      \cline{1-1}\cline{3-12}
      \texttt{ASGCN-RoBERTa}~\citet{DaiYSLQ21}               &                                                              & $83.33$                                  & $80.32$                                    & $86.87$                                    & $80.59$                                    & ---                               & ---                    & ---                    & ---                    & ---                    & ---                    \\
      \texttt{RGAT-RoBERTa}~\citet{DaiYSLQ21}                &                                                              & $83.33$                                  & $79.95$                                    & $87.52$                                    & $81.29$                                    & ---                               & ---                    & ---                    & ---                    & ---                    & ---                    \\
      \texttt{PWCN-RoBERTa}~\citet{DaiYSLQ21}                &                                                              & $84.01$                                  & $81.08$                                    & $87.35$                                    & $80.85$                                    & ---                               & ---                    & ---                    & ---                    & ---                    & ---                    \\
      \texttt{SARL-RoBERTa}$^{\dagger}$~\cite{WangSLZC21}    &                                                              & $85.42$                                  & $82.97$                                    & $88.21$                                    & $82.44$                                    & $88.19$                           & $73.83$                & $94.62$                & $81.92$                & ---                    & ---                    \\
      \texttt{RoBERTa}~\cite{LiuOGDJCLLZS19}$^{\ddagger}$    &                                                              & $82.76(0.63)$                            & $79.73(0.77)$                              & $87.77(1.61)$                              & $82.10(2.01)$                              & $78.06(0.55)$                     & $62.41(0.89)$          & $93.01(0.19)$          & $80.88(0.27)$          & $83.83(0.49)$          & $83.29(0.50)$          \\
      \cline{1-1}\cline{3-12}
      \texttt{DeBERTa}~\cite{HeGC21}$^{\ddagger}$            &                                                              & $82.76(0.31)$                            & $79.45(0.60)$                              & $88.66(0.35)$                              & $83.06(0.29)$                              & $87.50(0.28)$                     & $73.76(0.36)$          & $86.57(0.78)$          & $73.59(0.95)$          & $83.06(1.24)$          & $82.52(1.25)$          \\
      % \texttt{ASGCN-DeBERTa}~\citet{DaiYSLQ21}               &                                                              & $$                                  & $$                                    & $$                                    & $$                                    & ---                               & ---                    & ---                    & ---                    & ---                    & ---                    \\
      \texttt{SARL-DeBERTa}$^{\ddagger}$~\cite{WangSLZC21}   &                                                              & $83.32(0.42)$                            & $79.95(0.51)$                              & $86.69(0.27)$                              & $78.91(0.33)$                              & $86.53(0.19)$                     & $69.73(0.28)$          & $93.31(0.19)$          & $80.13(0.28)$          & $82.03(0.57)$          & $81.84(0.28)$          \\
      \hline
      \hline
      \texttt{\ourp-BERT}                                    & \multirow{14}[8]{*}{\begin{sideways}\our\end{sideways}}      & $81.35(0.63)$                            & $77.79(0.48)$                              & $87.23(0.22)$                              & $81.06(0.67)$                              & $84.07(0.78)$                     & $70.62(0.68)$          & $91.74(0.32)$          & $78.25(0.88)$          & $83.13(0.30)$          & $82.53(0.44)$          \\
      \texttt{\ourt-BERT}                                    &                                                              & $81.35(0.39)$                            & $78.43(0.52)$                              & $87.32(0.22)$                              & $81.86(0.20)$                              & $84.93(0.59)$                     & $73.01(0.79)$          & $91.42(0.45)$          & $77.50(0.86)$          & $83.51(0.26)$          & $82.90(0.28)$          \\
      \texttt{\ours-BERT}                                    &                                                              & $81.03(0.31)$                            & $77.45(0.37)$                              & $87.41(0.40)$                              & $81.52(0.49)$                              & $84.22(1.03)$                     & $71.98(0.85)$          & $91.58(0.54)$          & $77.54(0.71)$          & $83.23(0.56)$          & $82.68(0.52)$          \\
      \texttt{\oure-BERT}                                    &                                                              & $81.03(0.31)$                            & $77.45(0.37)$                              & $87.41(0.40)$                              & $81.52(0.49)$                              & $85.56(0.41)$                     & $73.79(0.57)$          & $92.20(0.63)$          & $78.49(0.65)$          & $83.23(0.56)$          & $82.68(0.52)$          \\
      % \texttt{\oure-BERT}
      \cline{1-1}\cline{3-12}
      \texttt{\ourp-RoBERTa}                                 &                                                              & $83.39(0.35)$                            & $80.47(0.44)$                              & $88.04(0.62)$                              & $82.96(0.48)$                              & $87.01(0.18)$                     & $73.71(0.31)$          & $90.31(0.94)$          & $76.17(1.48)$          & $83.37(0.31)$          & $83.78(0.29)$          \\
      \texttt{\ourt-RoBERTa}                                 &                                                              & $83.44(0.56)$                            & $80.47(0.71)$                              & $88.30(0.37)$                              & $83.09(0.45)$                              & $86.64(0.57)$                     & $72.24(0.79)$          & $94.22(0.71)$          & $83.41(1.45)$          & $83.31(0.41)$          & $84.60(0.22)$          \\
      \texttt{\ours-RoBERTa}                                 &                                                              & $83.23(0.44)$                            & $80.30(0.68)$                              & $88.48(0.52)$                              & $83.81(0.62)$                              & $88.31(0.47)$                     & $76.23(0.81)$          & $93.65(0.89)$          & $81.82(1.71)$          & $83.58(0.39)$          & $83.78(0.24)$          \\
      \texttt{\oure-RoBERTa}                                 &                                                              & $84.12(0.27)$                            & $80.90(0.51)$                              & $89.11(0.38)$                              & $83.98(0.69)$                              & $88.39(0.53)$                     & $76.19(0.68)$          & $94.15(0.64)$          & $82.18(1.38)$          & $85.48(0.29)$          & $85.02(0.17)$          \\
      \cline{1-1}\cline{3-12}
      \texttt{\ourp-DeBERTa}                                 &                                                              & $84.33(0.55)$                            & $81.46(0.77)$                              & $89.91(0.09)$                              & $84.90(0.45)$                              & $89.05(0.28)$                     & $77.14(0.37)$          & $93.49(0.43)$          & $81.44(0.53)$          & $83.91(0.31)$          & $83.31(0.21)$          \\
      \texttt{\ourt-DeBERTa}                                 &                                                              & $84.80(0.39)$                            & $82.00(0.43)$                              & $89.91(0.40)$                              & $85.05(0.85)$                              & $89.61(0.72)$                     & $79.17(0.12)$          & $93.65(0.39)$          & $81.53(0.51)$          & $84.28(0.32)$          & $83.70(0.47)$          \\
      \texttt{\ours-DeBERTa}                                 &                                                              & $84.17(0.08)$                            & $81.23(0.27)$                              & $89.64(0.66)$                              & $84.53(0.79)$                              & $89.42(0.38)$                     & $77.29(0.62)$          & $94.14(0.11)$          & $81.61(0.81)$          & $83.61(0.30)$          & $83.07(0.28)$          \\
      \texttt{\oure-DeBERTa}                                 &                                                              & $84.80(0.31)$                            & $82.09(0.31)$                              & $91.43(0.28)$                              & $86.85(0.19)$                              & $89.47(0.59)$                     & $77.84(0.40)$          & $94.47(0.37)$          & $82.39(0.27)$          & $85.85(0.18)$          & $85.29(0.37)$          \\
      \cline{1-1}\cline{3-12}
      \texttt{\ourpx-DeBERTa}                                &                                                              & $86.00(0.07)$                            & $83.10(0.30)$                              & $90.27(0.61)$                              & $85.51(0.48)$                              & $89.98(0.11)$                     & $78.26(0.98)$          & $95.11(0.69)$          & $84.68(0.21)$          & $82.78(0.96)$          & $81.99(0.86)$          \\
      \texttt{\ourtx-DeBERTa}                                &                                                              & $86.31(0.20)$                            & $83.93(0.27)$                              & $90.86(0.18)$                              & $86.26(0.22)$                            & $91.09(0.22)$                     & $81.22(0.34)$          & $94.71(0.56)$          & $84.34(0.38)$          & $84.21(0.42)$          & $83.72(0.46)$          \\
      \texttt{\oursx-DeBERTa}                                &                                                              & $86.21(0.52)$                            & $83.97(0.64)$                              & $90.33(0.37)$                              & $85.55(0.46)$                              & $90.63(0.17)$                     & $80.24(0.33)$          & $94.54(0.84)$          & $83.50(0.73)$          & $84.68(0.67)$          & $84.12(0.64)$          \\
      \texttt{\ourex-DeBERTa}                                &                                                              & $\mathbf{86.46 (0.38)}$                  & $\mathbf{84.41(0.39)}$                     & $\mathbf{90.98(0.28)}$                     & $\mathbf{87.02(0.42)}$                     & $\mathbf{91.85(0.27)}$            & $\mathbf{81.29(0.51)}$ & $\mathbf{95.61(0.64)}$ & $\mathbf{84.87(0.71)}$ & $\mathbf{86.38(0.29)}$ & $\mathbf{85.97(0.18)}$ \\
      \hline
    \end{tabular}%
  }
  \label{tab:main}%
  \vspace{-10pt}
\end{table*}%

\subsubsection{Cluster Prediction Performance}

We utilize \our\ to classify aspect sentiments and aggregate the sentiment clusters.
The cluster prediction performance in \pref{tab:cluster_prediction} shows that our models consistently outperform the baseline models on all datasets. The performance of \our\ is dependent on the base model.
It is observed that the sentiment clusters predicted by \our\ are very close to the ground truth, which demonstrates the effectiveness of our models in modeling sentiment coherency.
The small clusters (e.g., clusters containing $1$ or $2$ aspects) are more easy to predict, while the large clusters (e.g., $\geq 3$) are more difficult to predict.

\subsubsection{Sentiment classification performance}

When it comes to sentiment classification performance, the results in \pref{tab:main} clearly demonstrate the superiority of our models over significant baselines, particularly in the case of the \oure\ model.
The experimental results are as expected and show the proficiency of \our.

One of the primary concerns associated with \our\ is its occasional inability to outperform certain baselines based on the BERT model. We attribute this observation to two main reasons.
Firstly, \our\ is a quite simple mechanism and relies on relatively basic aspect features to construct sentiment aggregation windows, which may not be as competitive as state-of-the-art methods that employ more complex features.
Secondly, the current sentiment aggregation window, although intuitive, may not be perfect and could potentially lead to the loss of some sentiment information.
Nevertheless, the performance of the three \our\ variants may not consistently surpass some baselines, our models offer notable advantages in terms of efficiency and ease of integration with existing models.
With the improvement in the base model's performance (e.g., \texttt{DeBERTa}, \texttt{DeBERTa-Large}), \our\ achieves impressive results across all datasets.
Furthermore, it's worth noting that methods such as \texttt{ASGCN-RoBERTa}, \texttt{RGAT-RoBERTa}, and \texttt{PWCN-RoBERTa}, while showing promising improvements, come at the cost of significantly higher resource requirements compared to other models.

In summary, \our\ presents a compelling choice for a trade-off between performance and resource efficiency with the potential to be integrated into existing models with minimal effort.

\subsection{Practice in Adversarial Defense}
\begin{table}[htbp]
  \centering
  \caption{Performance comparison of different models for adversarial defense on the \texttt{Laptop14-ARTS} and \texttt{Restaurant14-ARTS} datasets. The adversarial datasets are from \citet{XingJJWZH20}.}

  \resizebox{\linewidth}{!}{
    \begin{tabular}{l|c|c|c|c}
      \hline
      \multirow{2}[1]{*}{Model} & \multicolumn{2}{c|}{\texttt{Laptop14-ARTS}} & \multicolumn{2}{c}{\texttt{Restaurant14-ARTS}}                                       \\
      \cline{2-5}               & \texttt{Acc}                             & \texttt{F1}                              & \texttt{Acc}     & \texttt{F1}      \\
      \hline
      \texttt{BERT}             & $63.98$                                  & $56.11$                                  & $72.01$          & $65.62$          \\
      \texttt{DeBERTa}          & $67.71$                                  & $65.60$                                  & $74.97$          & $66.48$          \\
      % \texttt{ASGCN-BERT-Base}    & $68.38$                                           & $61.91$                                  & $74.77$          & $67.40$          \\
      % \texttt{ASGCN-DeBERTa-Base} & $70.12$                                           & $63.54$                                  & $76.55$          & $68.59$          \\
      \hline
      \texttt{\ourp-BERT}       & $72.31$                                  & $68.94$                                  & $78.06$          & $70.23$          \\
      \texttt{\ourt-BERT}       & $72.12$                                  & $68.05$                                  & $77.57$          & $70.72$          \\
      \texttt{\ours-BERT}       & $70.88$                                  & $65.98$                                  & $77.99$          & $71.01$          \\
      \texttt{\oure-BERT}       & $\mathbf{74.32}$                         & $\mathbf{69.57}$                         & $\mathbf{78.41}$ & $\mathbf{72.04}$ \\
      \hline
      \texttt{\ourp-DeBERTa}    & $73.34$                                  & $68.46$                                  & $81.19$          & $72.54$          \\
      \texttt{\ourt-DeBERTa}    & $73.58$                                  & $69.28$                                  & $80.31$          & $71.37$          \\
      \texttt{\ours-DeBERTa}    & $72.31$                                  & $67.03$                                  & $79.13$          & $71.82$          \\
      \texttt{\oure-DeBERTa}    & $\mathbf{74.47}$                         & $\mathbf{69.79}$                         & $\mathbf{81.55}$ & $\mathbf{72.95}$ \\
      \hline
    \end{tabular}%
  }
  \label{tab:adversarial}%
  \vspace{-10pt}
\end{table}%

Recent works have highlighted the threat of textual adversarial attacks~\cite{XingJJWZH20} as critical threats.
In this section, we embark on a pioneering exploration of \our's capabilities, focusing on its ability to defend against adversarial attacks in ABSC.
To evaluate the robustness of \our\ in the face of these attacks, we employ existing adversarial attack datasets, specifically \texttt{Laptop14-ARTS} and \texttt{Restaurant14-ARTS}.

The results presented in \pref{tab:adversarial} serve as a testament to the superior performance of our models when compared to the baseline models, i.e., \texttt{BERT} and \texttt{DeBERTa}.
Notably, when considering the DeBERTa-based models, \texttt{\ourp-DeBERTa}, \texttt{\ourt-DeBERTa}, and \texttt{\ours-DeBERTa} consistently outperform the baselines, underscoring the robustness of \our\ in defend against adversarial attack.

\subsection{Ablation Study}
\label{sec:ablation}

\begin{figure*}[htbp]
  \centering
  \includegraphics[width=.9\linewidth]{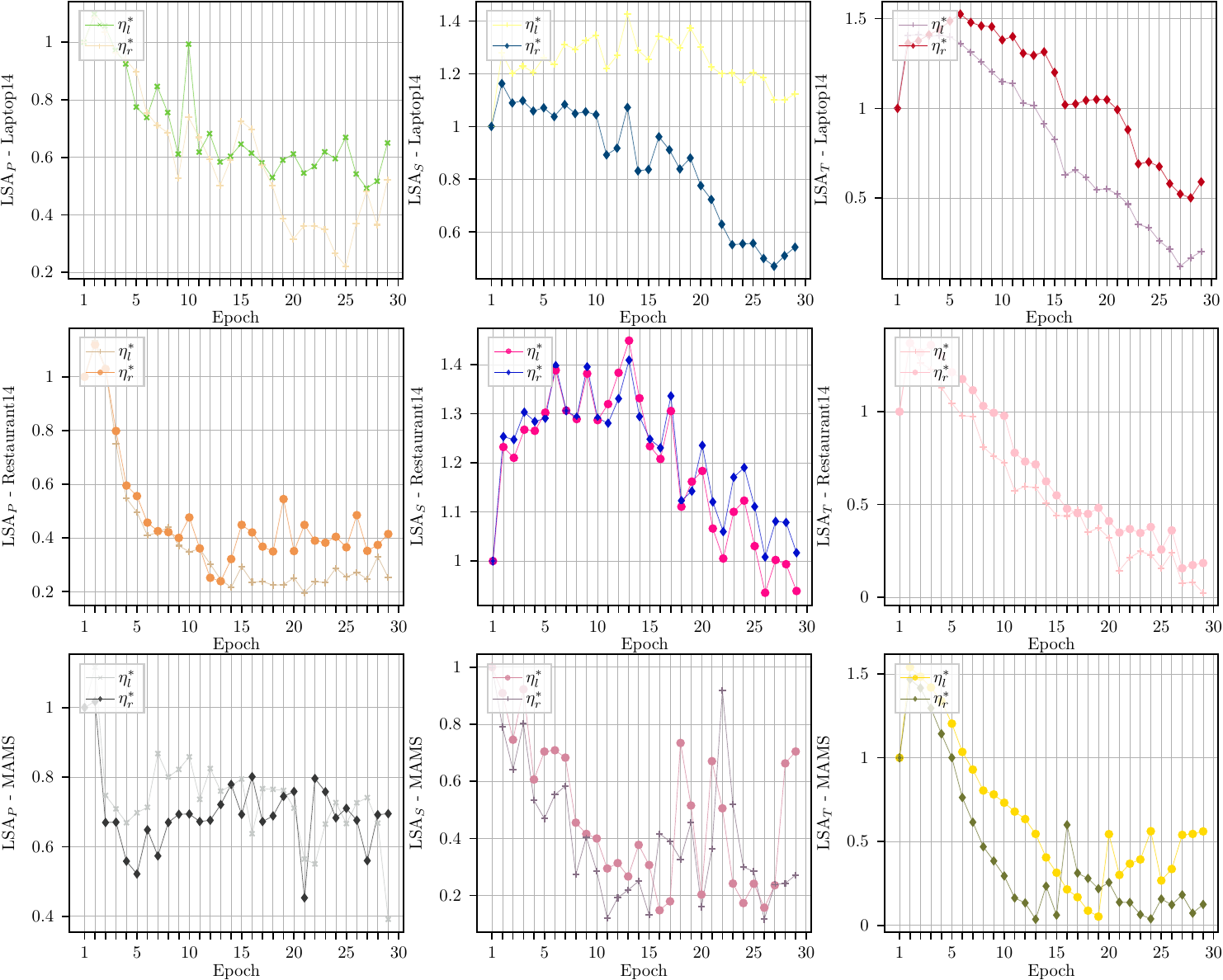}
  \caption{Trajectory visualization of learnable weights in gradient-based sentiment aggregation window optimization.}
  \label{fig:rq2dynamic_eta}
  \vspace{-10pt}
\end{figure*}

In this section, we study how gradient-based aggregation window optimization influences \our.
We begin by presenting the trajectory of $\eta_{l}^{*}$ and $\eta_{r}^{*}$ during the training process, as depicted in \pref{fig:rq2dynamic_eta}, which illustrates how \our\ dynamically constructs the optimal window.
This observation suggests that the model initially prioritizes the side aspects during early training stages, gradually shifting focus towards the central aspects.
To further investigate the impact of gradient-based aggregation window optimization, we conduct a comparative analysis by evaluating \our's performance with and two ablated models without DWA.
Specifically, we assess the model's performance when employing fixed static weights $\eta_{l}$ and $\eta_{r}$ to create sentiment aggregation windows, as opposed to the DWA.
The experimental results provided in \pref{fig:static_window} demonstrate a consistent performance drop when DWA is omitted.
In most scenarios, we observe a modest yet notable improvement of approximately 0.2\% to 0.5\% when DWA is incorporated into our model.
We also present the experimental results for an ablated version of \our\ featuring a simplified sentiment aggregation window in \pref{tab:simplified_window}.
This comparison underscores the superior performance of \our\ with DWA over its simplified counterpart.
Consequently, we can conclude that gradient-based aggregation window optimization proves effective in facilitating implicit sentiment learning.

\subsection{Case Study}

In this section, we delve into a case study to validate the capability of our model in learning local sentiment coherency.
We present a series of examples in \pref{tab:case_study}, which showcase instances where \our\ excels in identifying aspect sentiment coherency.

\begin{table}[htbp]
  \centering
  \caption{The examples for aspect sentiment coherency found by \our. The target aspects are denoted in \textbf{bold} and the \underline{underlined words} indicates the aspects with coherent sentiments. ``Pos'', ``Neg'' and ``Neu'' represent positive, negative and neutral, respectively. }
  \resizebox{\linewidth}{!}{

    \begin{tabular}{c|c|l|c|c}
      \hline
      \texttt{No.}            & \texttt{Domain}                & \multicolumn{1}{c|}{\texttt{Examples}}                  & \texttt{Model}                             & \texttt{Prediction}                                                                        \\
      \hline
      \multirow{4}[2]{*}{$1$} & \multirow{4}[2]{*}{Restaurant} & Not only was the \textit{food} \underline{outstanding}, & \multirow{2}[1]{*}{\ourp-\texttt{BERT}}    & \multirow{2}[1]{*}{Pos(Pos)~\textcolor{green}{\cmark}, Pos(Pos)~\textcolor{green}{\cmark}} \\
                              &                                & but also the \textbf{coffee} and \textbf{juice}!        &                                            &                                                                                            \\
      \cline{3-5}             &                                & Not only was the \textit{food} \underline{terrible},    & \multirow{2}[1]{*}{\ourp-\texttt{BERT}}    & \multirow{2}[1]{*}{Neg(Neg)~\textcolor{green}{\cmark}, Neu(Neg)~\textcolor{red}{\xmark}}   \\
                              &                                & but also the \textbf{coffee} and \textbf{juice}!        &                                            &                                                                                            \\
      \hline
      \multirow{4}[2]{*}{$2$} & \multirow{4}[2]{*}{Restaurant} & The \textit{servers} always \underline{surprise} us     & \multirow{2}[1]{*}{\ours-\texttt{BERT}}    & \multirow{2}[1]{*}{Pos(Pos)~\textcolor{green}{\cmark}}                                     \\
                              &                                & with a different \textbf{starter}.                      &                                            &                                                                                            \\
      \cline{3-5}             &                                & The \textit{servers} always \underline{temporize} us    & \multirow{2}[1]{*}{\ours-\texttt{BERT}}    & \multirow{2}[1]{*}{Neg(Neg)~\textcolor{green}{\cmark}}                                     \\
                              &                                & with a different \textbf{starter}.                      &                                            &                                                                                            \\
      \hline
      \multirow{4}[2]{*}{$3$} & \multirow{4}[2]{*}{TV}         & The speakers of this TV is \underline{great}!           & \multirow{2}[1]{*}{\ourt-\texttt{DeBERTa}} & \multirow{2}[1]{*}{Pos(Pos)~\textcolor{green}{\cmark}}                                     \\
                              &                                & Just like its \textbf{screen}.                          &                                            &                                                                                            \\
      \cline{3-5}             &                                & The speakers of this TV \underline{sucks}!              & \multirow{2}[1]{*}{\ourt-\texttt{DeBERTa}} & \multirow{2}[1]{*}{Neg(Neg)~\textcolor{green}{\cmark}}                                     \\
                              &                                & Just like its \textbf{screen}.                          &                                            &                                                                                            \\
      \hline
      \multirow{4}[2]{*}{$4$} & \multirow{4}[2]{*}{Camera}     & If you are worried about \textbf{usability},            & \multirow{2}[1]{*}{\texttt{DeBERTa}}       & \multirow{2}[1]{*}{Neu(Pos)~\textcolor{red}{\xmark}}                                       \\
                              &                                & think about \underline{the} \textit{quality} !          &                                            &                                                                                            \\
      \cline{3-5}             &                                & If you are worried about \textbf{usability},            & \multirow{2}[1]{*}{\texttt{DeBERTa}}       & \multirow{2}[1]{*}{Pos(Pos)~\textcolor{green}{\cmark}}                                     \\
                              &                                & think about \underline{it good} \textit{quality} !      &                                            &                                                                                            \\
      \hline
    \end{tabular}%
  }
  \label{tab:case_study}%
  \vspace{-10pt}
\end{table}%
These examples offer compelling evidence of the effectiveness of our model, as compared to a baseline model (\texttt{DeBERTa}).
For instance, in example \#4, the \texttt{DeBERTa} model produces two inference errors in recognizing coherent sentiments, while all our model variants based on the \texttt{DeBERTa} model yield correct results.
Furthermore, \ourp, \ourt, and \ours models demonstrate remarkable robustness in handling perturbed examples that involve local sentiment coherency.
While it is challenging to present a comprehensive list of sentiment cluster prediction examples, the consistent observations obtained in these experiments align with those in \pref{tab:case_study}.
Based on these experimental results, we confidently assert the model's proficiency in learning sentiment coherency within ABSC.

\section{Discussions}

\subsection{How can \our\ help to existing methods?}
The primary function of \our\ lies in aggregating aspect features based on local sentiment coherency. 
Thanks to its straightforward implementation, integrating \our\ into existing models is a seamless process. 
In practice, once aspect features have been extracted using any existing methods, \our\ can be effortlessly applied to extract aspect sentiment clusters, enhancing the overall performance of aspect sentiment classification.

A simple yet effective way to incorporate \our\ into existing models involves removing their output layer and passing the learned feature representations of adjacent aspects to \our. 
Subsequently, \our\ can construct the sentiment aggregation window and derive the weights for each aspect feature using the Differential Weighted Aggregation (DWA) method. 
% In our experiments, we have demonstrated that the integration of \our\ consistently enhances the performance of all backbone models.

\subsection{How does \our\ works on adverse sentiment aggregation?}

In this section, we justify why \our\ works for adjacent but inconsistent sentiment.
It is intuitively that not all aspect sentiments in adjacent positions are similar but sometimes be opposite.
However, \our\ learns to discriminate whether they share similar sentiments based on the training data.
If no local sentiment coherency is detected, \our\ learns a weight close to $0$ to the feature of adjacent aspects in the DWA.

We have conducted experiments on a sub-dataset extracted from the MAMS dataset that only includes both marginal aspects in clusters, denoted as \texttt{Margin} dataset.
We evaluate the sentiment prediction accuracy of aspects near inconsistent sentiment clusters.
The results are available in \pref{tab:margin}, and the performance of classifying margin aspects is still comparable to global performance in \pref{tab:main}, 
indicating that differentiated weighting for \our\ effectively mitigates the challenge of adverse sentiment aggregation.

\begin{table}[htbp]
    \centering
    \caption{The performance of sentiment predictions for margin aspects in various models on the \texttt{MAMS} dataset.}
    \resizebox{.8\linewidth}{!}{
        \begin{tabular}{l|c|c|c|c}
            \hline
            \multirow{2}{*}{\textbf{Model}} & \multicolumn{2}{c|}{\texttt{Margin}} & \multicolumn{2}{c}{\texttt{MAMS}}                                       \\
            \cline{2-5}
                                            & \texttt{Acc}                           & \texttt{F1}                       & \texttt{Acc}     & \texttt{F1}      \\
            \hline
            \ourp-\texttt{DeBERTa}          & $83.49$                                & $82.71$                           & $\mathbf{83.91}$ & $\mathbf{83.31}$ \\
            \ourt-\texttt{DeBERTa}          & $82.58$                                & $81.79$                           & $\mathbf{84.28}$ & $\mathbf{83.70}$ \\
            \ours-\texttt{DeBERTa}          & $\mathbf{83.87}$                       & $\mathbf{83.11}$                  & $83.61$          & $83.07$          \\
            \hline
        \end{tabular}
    }
    \label{tab:margin}
\end{table}

\section{Related Works}
The related works in this field can be broadly divided into three categories: sentiment dependency-based methods, sentiment coherency modeling, and implicit sentiment learning.

% \subsection{Sentiment Dependency-Based Methods}

Although sentiment coherency is prevalent in ABSC, it has received limited attention in recent years. 
However, the progress of sentiment dependency-based methods, such as the work by \citet{ZhangLS19,ZhouHHH20,TianCS21,LiCFMWH20,DaiYSLQ21}, has contributed to the improvement of coherent sentiment learning. 
These studies explored the effectiveness of syntax information in ABSC, which mitigates issues related to sentiment coherency extraction. 

For refining syntax structure quality in sentiment dependency learning, \citet{TianCS21} employ type-aware GCN to distinguish different relations in the graph, achieving promising results. 
Similarly, \citet{LiCFMWH20} propose \texttt{SynGCN} and \texttt{SemGCN} for different dependency information. 
\texttt{TGCN} model alleviates dependency parsing errors and shows significant improvement compared to previous GCN-based models. 
Despite the aforementioned advances, transferring the new techniques proposed in these studies is not straightforward. 
\citet{DaiYSLQ21} propose employing the pre-trained RoBERTa model to induce trees for ABSC, effectively solving the node alignment problem. However, the efficiency of inducing trees needs improvement.

% \subsection{Sentiment Coherency Modeling and Implicit Sentiment Learning}

Compared to coarse-grained implicit sentiment research~\cite{KauterBH15,ZhouWZH21,LiaoWCWZ22,ZhuangLLHC22}, the aspect's implicit sentiment learning in ABSC remains challenging. 
\our\ leverages coherency to aggregate implicit sentiments efficiently. Some researchers have formulated tasks aimed at modeling implicit sentiments and opinions. 
For instance, \citet{CaiXY20} proposed a quadruple extraction task (aspect, category, opinion, and sentiment), while \citet{Murtadha2022PWSZL} proposed a unified framework that crafts auxiliary sentences to aid implicit aspect extraction and sentiment analysis. 
In contrast to these works, \our\ sidesteps the efficiency bottleneck of syntax modeling by eliminating structure information and proves to be adaptable to existing methods as it is a transferable paradigm independent of base models.
\citet{LiZZZW21} presents a supervised contrastive pre-training mechanism to align the representation of implicit sentiment and explicit sentiment. 
However, it relies on fine-tuning a large-scale sentiment-annotated corpus from in-domain language resources, which may be resource-intensive and inefficient.

\section{Conclusion}
Aspect sentiment coherency has been overlooked in existing studies. 
We introduced the concept of \our, a novel approach that brings the nuance of local sentiment coherency into the foreground of ABSC. 
\our\ achieves state-of-the-art performance when combined with various aspect-specific features, especially based on the \texttt{DeBERTa} models. 
Furthermore, we also introduce a practice of \our\ in the realm of adversarial defense. 
We hope that our work will inspire further research into sentiment coherency modeling in the future.

\section{Limitations}

Although \our\ achieves impressive performance for multiple-aspects situations, e.g., \texttt{SemEval-$2014$} datasets. 
However, while being applied in mono aspect situations, such as the \texttt{Twitter} dataset, \our\ degenerates to be equivalent to a prototype model, e.g., the local context focus model.

Another limitation is that \our\ is a quite simple mechanism and relies on relatively basic aspect features to construct sentiment aggregation windows, which may not be as competitive as state-of-the-art methods that employ more complex features. 
Besides, the current sentiment aggregation window is intuitive but may not be perfect and could potentially lead to the loss of some sentiment information. 
In the future, we will explore more advanced sentiment aggregation windows to improve the performance of \our.

\section*{Acknowledgments}
This work was supported in part by the UKRI Future Leaders Fellowship under Grant MR/S017062/1 and MR/X011135/1; in part by NSFC under Grant 62376056 and 62076056; in part by the Royal Society under Grant IES/R2/212077; in part by the EPSRC under Grant 2404317; in part by the Kan Tong Po Fellowship
(KTP$\backslash$R1$\backslash$231017); and in part by the Amazon Research Award and Alan Turing Fellowship.

\bibliography{ref}
\bibliographystyle{acl_natbib}
\appendix

\newpage

\section{Challenges of Aspect Sentiment Cluster Extraction}
\label{app:challenges}
The challenges of concurrent aspect sentiment cluster extraction can be summarized in the following three aspects:
\begin{itemize}
	\item \textbf{Data Annotation:} Currently, there is no existing aspect cluster dataset in the literature since addressing sentiment coherence is a novel topic. Re-annotating cluster data and labels presents a significant challenge, and modeling these clusters is notably more complex when contrasted with local sentiment coherence aggregation.
	\item \textbf{Data Insufficiency:} Even after completing the data re-annotation process, the clusters within the datasets might still be insufficient for effectively training the model.
	\item \textbf{Modeling Difficulty:} Cluster mining is a hard task compared to text classification, but it is worth studying in the near future.
\end{itemize}

\section{Implementation Details}
\label{app:implementation}

\subsection{Model Architecture}
\label{app:model_architecture}
\begin{figure}[htbp]
	\centering
	\includegraphics[width=0.9\columnwidth]{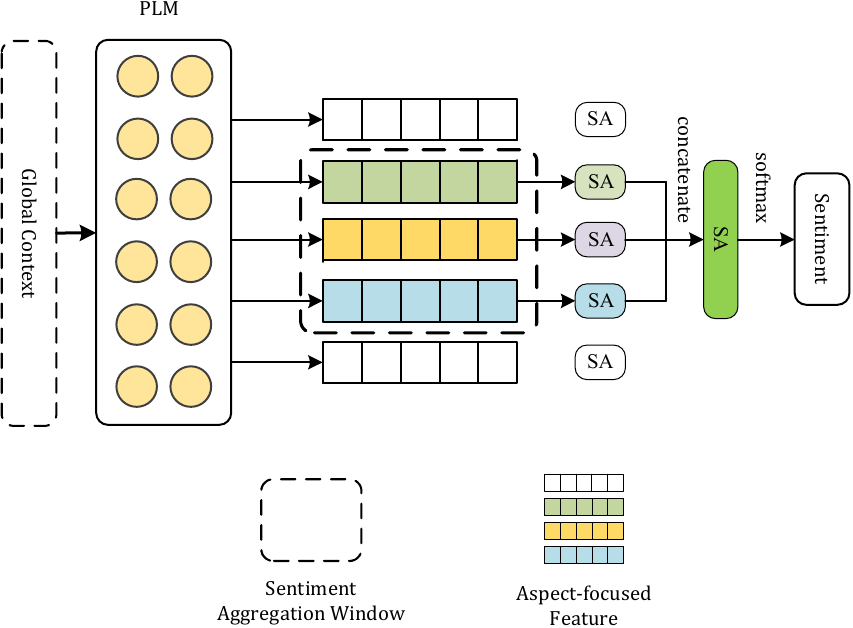}
	\caption{The local sentiment aggregation paradigm based on BERT-SPC, denoted as \ourp. ``SA'' indicates the self-attention encoder.}
	\label{fig:lsa-spc}
	\vspace{-10pt}
\end{figure}

We show the brief architecture of \ourp\ (based on the BERT-SPC input format) in \pref{fig:lsa-spc}. 
The input of \ourp\ is the same as BERT-SPC, which is a sequence of tokens with the aspect marked by the \texttt{[ASP]} token.

\section{Additional Experimental Results}
\label{app:additional_exp}

\subsection{Resource Occupation of \our}
The experiments are based on RTX2080 GPU, AMD R5-3600 CPU with PyTorch $1.9.0$. The original size of the \texttt{Laptop$14$} and \texttt{Restaurant14} datasets are $336kb$ and $492kb$, respectively.

\begin{table}[hp]
	\centering
	% \scriptsize
	\caption{The resources occupation of state-of-the-art ABSC models. ``\texttt{Proc.T.}'' and ``\texttt{Add.S.}'' indicate the dataset pre-processing time (sec.) and additional storage occupation (MB), respectively. ``$^*$'' represents non-syntax tree based models, and ``$^\dagger$'' indicates our models.}
	\resizebox{\linewidth}{!}{
		\begin{tabular}{l|c|c|c|c}
			\hline
			\multirow{2}[4]{*}{Model}         & \multicolumn{2}{c|}{\texttt{Laptop$14$}} & \multicolumn{2}{c}{\texttt{Restaurant14}}                                              \\
			\cline{2-5}                       & \texttt{Proc.T.}                         & \texttt{Add.S.}                           & \texttt{Proc.T.}         & \texttt{Add.S.} \\
			\hline
			\texttt{BERT-BASE} $^*$           & $1.62$                                   & $0$                                       & $3.17$                   & $0$             \\
			\texttt{LCF-BERT} $^*$            & $2.89$                                   & $0$                                       & $3.81$                   & $0$             \\
			\texttt{ASGCN-BERT}               & $13.29$                                  & $0.01$                                    & $0.02$                   & $9.4$           \\
			\texttt{RGAT-BERT}                & \textcolor{red}{$35.4k$}                 & $157.4$                                   & \textcolor{red}{$48.6k$} & $188$           \\
			\texttt{\ourt-BERT}$^*$$^\dagger$ & $3.16$                                   & 0                                         & $4.32$                   & $0$             \\
			\texttt{\ours-BERT}$^*$$^\dagger$ & $20.56$                                  & 0                                         & $30.23$                  & $0$             \\
			\texttt{\ourp-BERT}$^*$$^\dagger$ & $0.20$                                   & 0                                         & $0.32$                   & $0$             \\
			\hline
		\end{tabular}
	}
	\label{app:tab_rq4}
\end{table}

\subsection{Experiment of Static Weighted Sentiment Aggregation}
\label{app:static_window}

\begin{figure}[htbp]
	\centering
	\includegraphics[width=\columnwidth]{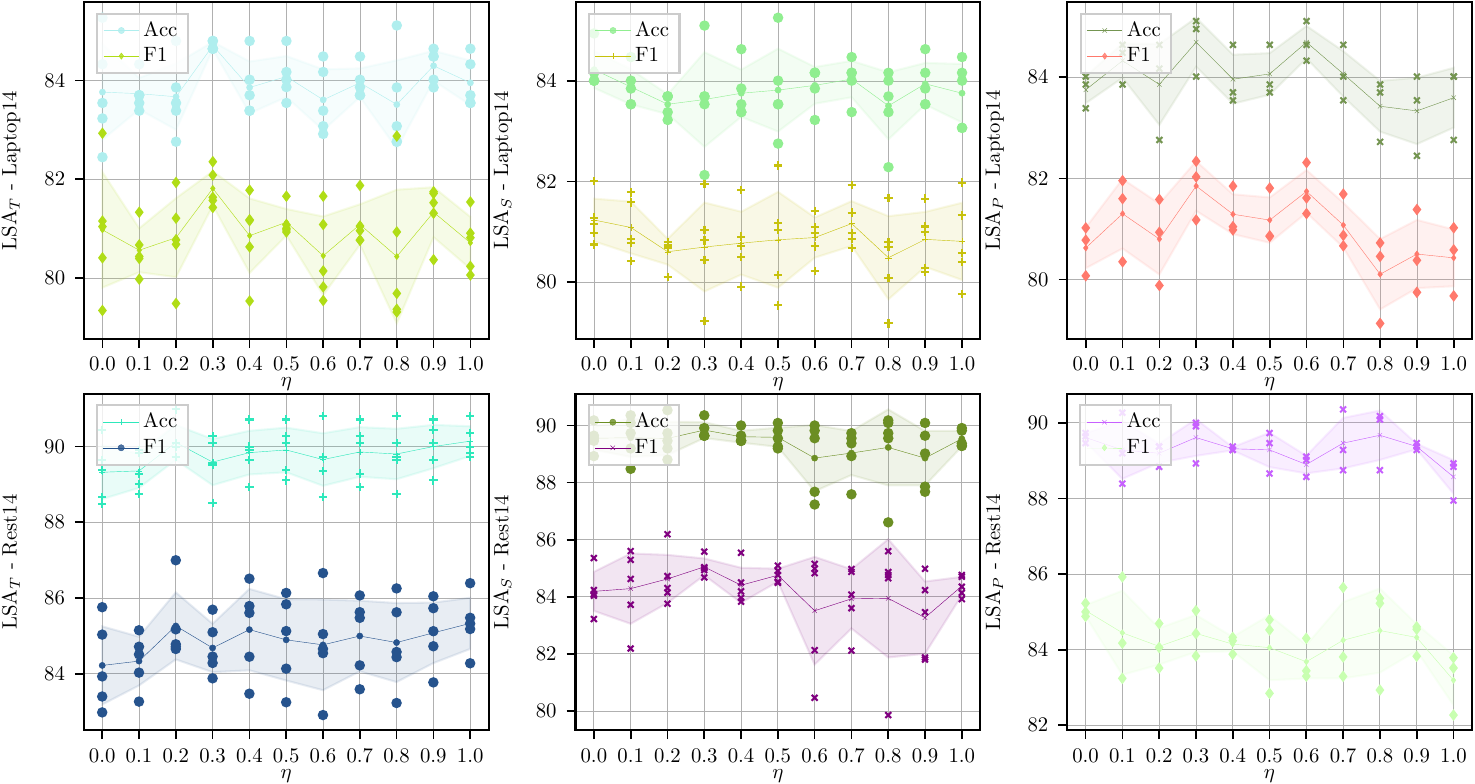}
	\caption{Visualization of performance under static differential weighting.}
	\label{fig:static_window}
\end{figure}

Besides the dynamic sentiment window differential weighting, we also try static weight to control the contribution of adjacent aspects' sentiment information. 
We first initialize $\eta_{l}$, $\eta \in [0, 1])$, for the left-adjacent aspects, while $\eta_{r}=1-\eta_{l}$. 
In this case, a greater $\eta_{l}$ means more importance of the left-adjacent aspect's feature and vice versa. 
However, it is difficult to search for the optimal static weights for many scenarios via gird search. 
We even found that the performance trajectory is non-convex while $\eta_{l} \in [0,1]$, indicating the $\eta_{l}$ on a dataset will be difficult to reuse on another dataset. 
\pref{fig:static_window} shows the performance curve of \our\ based on \texttt{DeBERTa} under different $\eta_{l}$.

In other words, static differential weighting is inefficient and unstable. We recommend applying an automatic weights search to find a better construction strategy for the sentiment window.

\subsection{Clarification of Hyper-parameter ``$k$'' Setting}
In this work, all experiments are implemented with $k=1$. The term ``$k=1$'' indicates that we only consider one-hop adjacent aspects for learning sentiment coherency. When $k = 2$, \our\ will consider five aspects in the sentiment aggregation windows. This setting performs well for handling sentiment clusters containing fewer than five aspects ($k = 2$). We did not conduct an ablation study of $k$ because the clusters in most datasets are not very large, and efficiency could be a problem. Below, we show the ratio of clusters with fewer than $5$ aspects versus those with $5$ or more aspects. It is observed that only a few sentiment clusters contain more than five aspects. Additionally, efficiency significantly decreases when the sentiment aggregation window increases to $5$ (i.e., $k=2$).
\begin{table}[htbp]
    \centering
    \caption{The proportion of aspect clusters with different sizes in different public ABSC datasets.}
    \resizebox{\linewidth}{!}{
        \begin{tabular}{c|c|c}
            \hline
            \multirow{2}{*}{\textbf{Dataset}} &\texttt{Cluster Size $ < 5$} & \texttt{Cluster Size $\geq 5$}                   \\
            \cline{2-3} & \textbf{Acc}     &     \textbf{Acc}     \\
            \hline
            \texttt{Laptop14}          & $79.30$             & $20.70$     \\
            \texttt{Restaurant14}          & $74.32$             & $25.68$ \\
            \texttt{Restaurant15}          & $81.28$             & $18.72$    \\
            \texttt{Restaurant16}          & $80.43$             & $19.57$ \\
            \texttt{MAMS}          & $88.84$             & $11.16$       \\
            \hline
        \end{tabular}
    }
    \label{tab:k}
\end{table}

\subsection{Experiment of Simplified Sentiment Aggregation Window}
\label{app:simplified_window}
To investigate the necessity of bidirectional aggregation, we assess the effectiveness of the streamlined aggregation window. 
We simply concatenate the left or right adjacent aspect's feature with the targeted aspect's feature and then change the output layer to accommodate the new feature dimension of the simplified aggregation window.

% Table generated by Excel2LaTeX from sheet 'Main'
\begin{table}[htbp]
	\centering
	% 	\footnotesize
	\scriptsize
	\caption{The average performance deviation of ablated \our\  baselines. ``LA'' and ``RA'' indicates the simplified aggregating window constructed only exploits the left-adjacent aspect or right-adjacent aspect, respectively.}
	\resizebox{\linewidth}{!}{
		\begin{tabular}{l|c|c|c|c}
			\hline
			\multirow{2}[2]{*}{\textbf{Model}} & \multicolumn{2}{c}{\texttt{Laptop$14$}} & \multicolumn{2}{c}{\texttt{Restaurant$14$}}                                                                         \\
			\cline{2-5}
			                                   & \multicolumn{1}{c|}{\texttt{Acc}}       & \multicolumn{1}{c|}{\texttt{F$1$}}          & \multicolumn{1}{c|}{\texttt{Acc}} & \multicolumn{1}{c}{\texttt{F$1$}} \\
			\hline
			\texttt{\ourp-DeBERTa}             & $\mathbf{84.33(0.37)}$                  & $\mathbf{81.46(0.52)}$                      & $\mathbf{89.91(0.33)}$            & $\mathbf{84.90(0.49)}$            \\
			-- w/ LA                           & $83.65(0.47)$                           & $80.48(0.62)$                               & $89.20(0.28)$                     & $84.26(0.31)$                     \\
			-- w/ RA                           & $83.86(1.25)$                           & $80.41(1.26)$                               & $88.57(0.65)$                     & $83.16(0.78)$                     \\
			\hline
			\texttt{\ourt-DeBERTa}             & $84.16(0.31)$                           & $81.40(0.55)$                               & $\mathbf{89.91(0.43)}$            & $\mathbf{84.96(0.40)}$            \\
			-- w/ LA                           & $84.08(1.25)$                           & $81.21(1.51)$                               & $89.55(0.62)$                     & $84.68(1.13)$                     \\
			-- w/ RA                           & $\mathbf{84.39(0.78)}$                  & $\mathbf{81.54(1.22)}$                      & $89.38(0.45)$                     & $83.99(0.68)$                     \\
			\hline
			\texttt{\ours-DeBERTa}             & $\mathbf{84.33(0.31)}$                  & $\mathbf{81.68(0.44)}$                      & $\mathbf{90.27(0.76)}$            & $\mathbf{85.78(0.56)}$            \\
			-- w/ LA                           & $83.57(1.10)$                           & $80.44(1.14)$                               & $89.29(0.89)$                     & $84.00(1.22)$                     \\
			-- w/ RA                           & $83.95(0.47)$                           & $80.89(0.88)$                               & $89.55(0.40)$                     & $84.26(0.39)$                     \\
			\hline
		\end{tabular}%
	}
	\label{tab:simplified_window}%
\end{table}%

\pref{tab:simplified_window} shows the experimental results. 
From the performance comparison of simplified aggregation, we observe that the full \our\ is optimal in most situations, despite the underlying PLM or training dataset. 
Moreover, to our surprise, \our\ with ``RA'' outperforms \our\ with ``LA'' in some situations. 

\subsection{Experiments on Twitter Dataset}
\label{sec:twitter}
The experimental results on the Twitter dataset reveal that the extended \ourx\ models, with \ourtx-DeBERTa demonstrating the best performance, 
effectively leverage local sentiment coherency to achieve competitive accuracy and F1 scores while maintaining consistent results across different runs.
\begin{table}[htbp]
	\centering
	\footnotesize
	\caption{The performance of \our\ models on the \texttt{Twitter} datasets, and the best results are heightened in \textbf{bold}. Numbers in parentheses denote IQR.}
	\resizebox{\linewidth}{!}{
		\begin{tabular}{l|c|c|c}
			\hline
			\multicolumn{1}{l}{\multirow{2}[1]{*}{\textbf{Model}}}      &                                                         & \multicolumn{2}{c}{\texttt{Twitter}}                          \\
			\cline{3-4}        \multicolumn{1}{c}{\multirow{2}[1]{*}{}} &                                                         & \texttt{Acc}                         & \texttt{F1}            \\
			\hline
			\texttt{\ourp-DeBERTa}                                      & \multirow{3}[1]{*}{\begin{sideways}\our\end{sideways}}
			                                                            & $76.91(0.36)$                                           & $75.90(0.41)$                                                 \\
			\texttt{\ourt-DeBERTa}                                      &                                                         & $76.61(0.20)$                        & $76.12(0.27)$          \\
			\texttt{\ours-DeBERTa}                                      &                                                         & $76.61(0.52)$                        & $75.84(0.64)$          \\
			\hline
			\texttt{\ourpx-DeBERTa}                                     & \multirow{3}[1]{*}{\begin{sideways}\ourx\end{sideways}}
			                                                            & $76.81(0.76)$                                           & $76.09(0.50)$                                                 \\
			\texttt{\ourtx-DeBERTa}                                     &                                                         & $\textbf{77.17(0.71)}$               & $\textbf{76.45(0.65)}$ \\
			\texttt{\oursx-DeBERTa}                                     &                                                         & $77.06(0.26)$                        & $76.23(0.29)$          \\
			\hline
		\end{tabular}%
	}
	\label{tab:twitter}%

\end{table}%

\end{document}